# Learning Periodic Human Behaviour Models from Sparse Data for Crowdsourcing Aid Delivery in Developing Countries

**James McInerney, Alex Rogers, Nicholas R. Jennings**
University of Southampton, Southampton, SO17 1BJ, UK
{jem1c10,acr,nrj}@ecs.soton.ac.uk

## Abstract

In many developing countries, half the population lives in rural locations, where access to essentials such as school materials, mosquito nets, and medical supplies is restricted. We propose an alternative method of distribution (to standard road delivery) in which the existing mobility habits of a local population are leveraged to deliver aid, which raises two technical challenges in the areas optimisation and learning. For optimisation, a standard Markov decision process applied to this problem is intractable, so we provide an exact formulation that takes advantage of the periodicities in human location behaviour. To learn such behaviour models from sparse data (i.e., cell tower observations), we develop a Bayesian model of human mobility. Using real cell tower data of the mobility behaviour of 50,000 individuals in Ivory Coast, we find that our model outperforms the state of the art approaches in mobility prediction by at least 25% (in held-out data likelihood). Furthermore, when incorporating mobility prediction with our MDP approach, we find a 81.3% reduction in total delivery time versus routine planning that minimises just the number of participants in the solution path.

## 1 INTRODUCTION

In many developing countries (e.g., Ivory Coast, Ghana, Liberia, Nigeria), half the population lives in rural locations [5], where accessibility to school materials, medical supplies, mosquito nets, and clothing is restricted. Distribution to these locations typically requires direct road transport, which is time consuming and requires bulk volume to be cost effective. In response to these limitations, distributed methods of aid distribution have emerged in recent years.

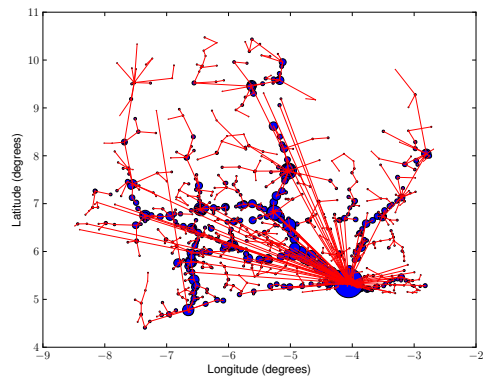

Figure 1: Minimum spanning tree between cell towers in Ivory Coast, where connections are defined by common visitors, and the size of node represents its betweenness centrality (i.e., the number of times the location appears in the shortest path for all possible delivery paths).

For example, Pack For a Purpose[1] is a non-profit organisation that asks tourists who already have a trip planned for one of 47 developing countries to bring small items (e.g., pencils, deflated soccer balls, stethoscopes) in their spare luggage capacity. Another scheme is Pelican Post[2], which asks donors to send books by post to developing countries. These are promising schemes. However, they fail during periods of conflict, (e.g., post-electoral violence in Ivory Coast in 2011) and are reliant on direct outsider support, when it is arguably preferable to empower local populations wherever possible.

In this work, we propose a new distribution method that uses the natural mobility of a local population to distribute physical packages from one location to another. In more detail, we wish to take advantage of the pre-existing mobility routines of a set of local participants by asking them to pick up a package from one exchange point (at a location that they normally visit, at a time that they normally visit it) and then drop it off at another exchange point (e.g., a lockbox or affiliate store) that is also part of their regular mobility. By chaining together the mobility of several participants, we may cover a large area, possibly a whole country, without having to deploy more expensive and time consuming infrastructure.

---
[1] http://www.packforapurpose.org
[2] http://www.pelican-post.org

While potentially appealing, this vision of crowdsourcing physical package delivery faces two signiÞcant technical barriers in optimisation and learning[3].

In optimisation, the possible delay between stages in the package's journey is unbounded, since the delay introduced by each participant is unknown and has no upper limit. This makes it infeasible to optimise the selection of participants and the package route (given a specific delivery problem specifying the start time, source location, and destination location), as delays propagate through the system [11]. In general, routing under delay uncertainty is a #P-hard problem to solve optimally [17]. Therefore, we formulate the decision problem in a way that takes advantage of the periodicities in human location behaviour to derive an exact solution.

In learning, the historical movements of individuals may be obtained from cell tower connections registered by mobile devices, which have widespread adoption across the developing world. However, such mobility data is sparse: it is limited in duration (i.e., we may only have a few week's worth of data from each participant) and, crucially, cell tower readings are taken only when a call or text message is exchanged from the phone, so there are large periods when no location of an individual is registered at all. Yet, existing methods for mobility prediction rely on large quantities (covering several weeks) of fairly continuous stream of location readings (either from GPS or constant cell tower monitoring) [23, 6, 14]. To overcome this, we develop a robust Bayesian model of individual mobility that can be learnt from cell tower records spanning only short periods of time with sporadic observability.

In more detail, we make the following three contributions:

- We advance the state of the art in route planning in delay networks by developing an approach that works well with the uncertainties caused by routine human behaviour. Specifically, we show that an exact and tractable solution is possible when using a mobility model belonging to the broad class of *temporal periodic* prediction models. Under this assumption, we show that we can formulate the problem as a Markov decision process (MDP) in which the number of states grows linearly in the number of locations, making the overall algorithm polynomial when using a standard MDP solving method (e.g., linear programming, policy iteration) [20]. Using our approach, simulations indicate that source-to-destination delivery time is reduced by an average of 81.3% compared to choosing the shortest path (which naïvely minimises the number of intermediate stages in the package's journey).

- To provide accurate transition probabilities to the MDP[4], we present a Bayesian nonparametric mixture model approach to learning the mobility behaviour of individuals from very sparse observations. We show how this model can be formulated as a series of Bernoulli trials and directly incorporated into the MDP. Using real cell tower data from 50,000 people in Ivory Coast (provided by Orange), we find at least a 25% improvement in held-out data likelihood when compared to two state-of-the-art approaches for human location behaviour prediction (a variable-order Markov model with prediction by partial matching [25] and a daily periodic finite mixture model [4])

- We use the Orange dataset to show that peer-to-peer package delivery is feasible under three key criteria. In particular, we show that the size of participant pool only needs to be of the order of several thousand to get at least an 80% coverage of the country (out of a total area 320,000 km$^2$). Furthermore, each solution path (i.e., chain of participants to deliver a package) is between 2-4 people. Finally, these requirements are only mildly worsened when considering only rural destinations for delivery.

The rest of the paper is structured as follows. First, in Section 2, we consider previous work related to the problem of learning human mobility patterns and optimising under uncertainty of human behaviour. In Section 3, we present our approach, starting with how we make optimal decisions with respect to the choice of participants and locations for any given delivery problem in Section 3.1. Then, in Section 3.2, we present a learning model that deals with sparse observations. In Section 4, we evaluate the feasibility of the scenario before evaluating our approach to learning and optimisation against several state of the art benchmarks. We draw conclusions and outline future work in Section 5.

## 2 RELATED WORK

The idea of distribution using the natural mobility of a group of people is a reoccurring theme in content distribution using mobile *ad-hoc* networks. For example, Keller et al. (2012) used physical bluetooth proximity data from the mobile phones of a group of people, to initiate exchanges of songs between individuals, but without considering prediction or multi-hop routes (i.e., going via one or more intermediaries) [10]. Cherubini et al. (2010) explored physical package peer-to-peer delivery, but only tested simple heuristics such as "transfer the package to someone who is, on average, closer to the target location than you" [3]. Vukadinovic et al. (2009) proposed a queuing model of the flow of pedestrian crowds to distribute content among mobile phones [27]. Now, all these works attempt to capture short term movements of individuals in crowds, which is a distinct and different problem to extracting *routine* mobility patterns. Specifically, in our work, there is a direct line of assumptions going from the raw historical data to decision-making about distribution (via learning and the formulation of transition probabilities in the MDP) that is not present in such work. A notable exception is by Liu and Wu (2011), who used class attendance data to model pairwise encounters between individuals for data transfer across an *ad hoc* network [13]. Like our work, they also take advantage of cyclic behaviour to find tractable routing solutions, however, their pairwise approach means that their algorithm scales $O(p^2)$ in the number ($p$) of participants in the network, while our approach scales polynomially only in the number of locations. In general, content distribution approaches often rely on the fact that content may be copied and can exist concurrently on multiple devices, making them less applicable for our routing problem.

---

[3]In addition to social issues such as trust (e.g., theft or loss) that we only consider briefly, in Section 4.4.

[4]N.B., the transition probabilities in the MDP are *not* the same as the transition probabilities of the mobility of any individual participant.

Another type of diffusion that attracts intense research interest is the study of the spread of infectious diseases. Epidemiologists look at the mobility dynamics of a population to identify source regions (from which disease is spread), and likely importation regions (to which disease is spread). For example, Wesolowski et al. (2012) used one year of cell phone data of millions of people to model the human movement between different regions in Nairobi [28]. They considered a graph in which the weight of the edges represents the quantity of people travelling between different locations. Hufnagel et al. (2004) considered a global model of human movement using passenger numbers for flights between the 500 largest airports in the world [9]. Such work is concerned with *aggregate* mobility, in contrast, we are interested in *individual* mobility, because, eventually, we need to ask specific people to contribute. Furthermore, we consider a full decision-making model, in addition to a purely descriptive model of human location behaviour.

The problem of robust route planning under uncertainty resembles the *Canadian traveller* problem (also known as the *bridge* problem) [18], in which the costs of the edges in a graph are random variables that are observed only as the nodes are visited. The name originates from the concept of a traveller who has to plan a journey between two locations, where the costs of outgoing edges are random variables that are only observed as a graph is traversed. This differs from our problem because the Canadian traveller assumes that path costs are independent of one another, while we have dependencies between costs as well, i.e., the delay outcome of an earlier stage in the chain affects the delay of later stages. An additional difference is that we observe the random variables, indicating delay between locations, only *after* the package has completed each intermediate step.

Learning routine mobility models has typically been a separate problem from optimisation. Approaches range from purely temporal ([15, 23, 24]), spatial ([7, 25]), to a combination of both ([6]). Existing datasets that are widely available have tended to contain approximately continuously recorded cell towers or GPS (e.g., the Reality Mining dataset recorded the cell tower every few minutes [6]; the Nokia dataset recorded GPS every few minutes also [12]). This has inspired many methods that work well on continuous location updates, but which do not perform as well as their headline accuracy (when predicting future location behaviour) on sparse data. We address this issue in our work. Given their ability to refine the model as more data arrives, nonparametric Bayesian methods are surprisingly rare in the literature on predicting human location behaviour. Chen et al. (2012) used a Gaussian process to model congestion on road networks, while Gao et al. (2012) used a hierarchical Pitman Yor process to model check-in behaviour on location-based social networks [2, 7].

Finally, crowdsourcing teams of participants who function as a chain to achieve a single goal resembles the idea behind the winning entries to the DARPA Red Balloon Challenge [19] and the Tag Challenge [21]. This work is primarily concerned with the problem of recruiting individuals and verifying their reports, which requires designing economic mechanisms. In this work, we assume recruitment can be done beforehand by an appropriate method (i.e., we do not address it here) but we do investigate *how many* participants are required for satisfactory delivery results.

## 3 DECISION-MAKING WITH UNCERTAIN HUMAN LOCATIONS

In this section, we present our approaches towards optimisation and learning with uncertain human behaviour in the package delivery scenario. Specifically, in Section 3.1, we show how it is possible to tractably find an exact optimal solution to routing under delay uncertainty, given a wide class of mobility model (which we define as *temporal periodic* models). In Section 3.2, we give more detail on our probabilistic mobility model that is designed to function well with sparse mobile phone datasets, and provides the predictions used in the optimisation.

### 3.1 THE OPTIMISATION PROBLEM

We formulate the optimisation problem sketched in Section 1 as an MDP, as this provides a principled way of making decisions under uncertainty. Decisions in this scenario must specify which participants to ask to pick up the package, from where they should pick it up, as well as the drop-off location. We assume the delay between pick-up and drop-off is outside the planner's control (so we treat it as a random variable here), and completely up to the participant who, when asked, does this according to his/her routine schedule.

In general, an MDP is defined as a tuple $(S, A, R, T)$ where $S$ is a set of states, $A$ is a set of available actions for each state, $R(s, a, s')$ is the function that specifies the cost of doing action $a \in A$ to get from state $s$ to $s'$, and $T(a, s, s')$ is the probability of getting from state $s$ to $s'$ when performing action $a$[5]. The solution to an MDP consists of an optimal policy, $q(s)$, that specifies the best action to perform for any given state $s$. Ancillary to this function is the value function, $G$, which gives the expected value for any state (given that the optimal action is performed). We consider each of $A$, $S$, $R$, and $T$ in turn.

### 3.1.1 Set of Actions $A$

We assume that the planner has no direct control over the delay (it is up to the participant's schedule) but we assume that we are guaranteed to eventually reach location $w$, when performing action $a$ (going to location $w$), and that the arrival time is revealed only after performing each action, resulting in a transition to state $(v, t_v)$, with unknown arrival time $t_v$. Given the one-to-one mapping of actions and locations (specifying the destination location) we treat locations as synonymous with actions.

### 3.1.2 Set of States $S$

We define the set of states $S$ in the MDP as the set of tuples describing the possible locations and times $(v, t_v)$ (respectively) of the package. This results in the set $S = \{(v, t_v) | v \in V, t_v = 1, 2, 3, ...\}$. We assume discrete time $t$ to capture the required detail in the scenario without the need for more complex continuous time reasoning. However, even in the discrete time case, we see that there is an unbounded number of states in $S$ because the delay in moving between locations is unbounded. This makes the standard MDP formulation intractable.

To overcome large state spaces, there are a few general approaches such as sampling methods or value approximation (in which values are computed from features of the states) [29]. One time-specific approach is to truncate the range of values for $t$ to find an approximation for the optimal policy [26]. However, the number of states grows as a factor of this truncation limit, so more exact approximations must be traded off with computation time.

---

[5]It is typical to include a time discount factor for future rewards in an MDP, however, this assumption makes less sense when utility is a function of delay. Therefore, we omit it in our model.

Instead, we find an exact solution under an additional assumption about the mobility model used to produce the probabilistic delays. Specifically, we show that for a large class of mobility models, namely *periodic temporal* models, the probability of delay, $pr(t_w - t_v | v, t_v, w)$ in going from state $(v, t_v)$ to $(w, t_w)$ is periodic in $t_v$. This results in an MDP with a linear number of states in the number of locations. This assumption is suitable for optimisation in delay networks, since it is precisely the periodic temporal class of mobility model that is most useful in predicting and planning several days in advance, since short term spatial correlations (e.g., a participant tends to go home after visiting the market, or always goes to the city centre after travelling along a particular road) do not have much effect beyond several hours. However, this assumption of temporal periodicity means that we cannot incorporate the most recent observations into the model, which may provide a benefit in optimising decisions to be made in the very near future. Under this assumption, we now establish linearity in the number of locations.

**Theorem 1.** Let $S$ be the set of states $\{(v, t_v) | v \in V, t_v = 1, 2, 3, ...\}$ in an MDP. If $pr(v|t_v)$ is a periodic function (defining $H$ as the number of possible values it can take) in discrete $t_v$ ($\forall v$), then the number of states is linear in the number of locations, i.e., $|S| = H |V|$.

*Proof*: Let $pr(v|t_v)$ be the probability that a given participant is at location $v$ at time $t_v$, obtained from a mobility model (which, we emphasise, describes individual behaviour and is distinct from the transition function $T$ of the MDP defined in Section 3.1.3). Since $t_v$ is discrete, we can repeat Bernoulli trials from the distribution $r_{d_v} \sim Bern(pr(v|t_v + d_v))$ for increasing $d_v = 1, 2, 3, ...$ until we get $r = 1$. This is a standard formulation (equivalent to repeated tosses of biased coins), with $pr(d_v|t_v) = pr(v|t_v + d_v) \prod_{d'_v=1}^{d_v-1} (1 - pr(v|t_v + d'_v))$. Since $pr(v|t_v)$ is periodic in $t_v$, with a maximum of $H$ distinct values, the probability of delay, $pr(w|t_v + d_v)$, *from* any next location $w$ (reachable from $v$) is also periodic for arbitrary delay $d_v$. Therefore, $pr((t_v + d_v) \mod H)$ is a sufficient statistic for $pr(d_w | t_v + d_v)$ (the probability of delay $d_w$ from $w$), clearly taking at most $H$ values. Using the Markov property of MDPs, only $H$ states are required for each location $v$ (for arbitrary $v$), resulting in $H |V|$ states overall. □

Unlike a truncation parameter, we can easily set $H$ for the specific application of the delay network that needs to be modelled, without bias (i.e., without underestimating the delay). For package delivery, we found it sufficient to set $H = 14$ per week, by considering the probability of a participant dropping off or picking up the package in slots of half a day. Therefore, the state space is now $S = \{(v, t) | v \in V, t \in [1, 14]\}$.

### 3.1.3 Cost Function $R$ and Transition Function $T$

The delay in going from location $v$ to location $w$ is the cost function $R(s, a, s')$, where $s = (v, t_v)$, $s' = (w, t_w)$, and $a$ is the action of routing the package to $w$. The MDP requires a single cost for each state $s$ and action $a$ pair (marginalising over the destination action), yet we have many participants who can potentially perform that action (i.e., who routinely visit both $v$ and $w$ locations). We define the *best person* as the one who minimises $p^* = \arg\min_i \{\mathbb{E}(d_{v,w}|t_v, i) + \sum_w c_w pr(t_w|t_v, i)|p_i \in P\}$, the cost of going from location $v$ to $w$ plus the expected cost of $c_w$ (the total cost at state $(w, t_w)$). The cost function $R$ is then the sum of delays for the best person to pick the package up at location $v$, and drop the package off at $w$:

$$R((v, t_v), w, (w, t_w))$$
$$= \mathbb{E}(d_v | t_v \bmod H) + \mathbb{E}(d_w | t_v + d_v \bmod H)$$
$$= \sum_{i=0}^{\infty} W_v^i (Hi + d_v) pr(d_v) \prod_{d'_v=1}^{d_v-1} (1 - pr(d'_v))$$
$$+ \sum_{i=0}^{\infty} W_w^i (Hi + d_w) pr(d_w) \prod_{d'_w=1}^{d_w-1} (1 - pr(d'_w))$$

(1)

where $W_v = \prod_{d'_v=1}^{H} (1 - pr(d'_v))$ and $W_w = \prod_{d'_w=1}^{H} (1 - pr(d'_w))$, with the respective interpretations being the probability of the participant *not* visiting the start and end locations (respectively) for an entire period. We now find the geometric sum:

$$R((v, t_v), w, (w, t_w)) =$$
$$\left( \frac{d_v}{1 - W_v} + \frac{W_v H}{(1 - W_v)^2} \right) pr(d_v | t_v) \prod_{d'_v=1}^{d_v-1} (1 - pr(d'_v | t_v))$$
$$+ \left( \frac{d_w}{1 - W_w} + \frac{W_w H}{(1 - W_w)^2} \right) pr(d_w | t_v + d_v) \cdot$$
$$\cdot \prod_{d'_w=1}^{d_w-1} (1 - pr(d'_w | t_v + d_v))$$

(2)

The transition function $T(a, s, s')$ may be found in a similar way, but by considering only whole multiples of the given delay:

$$T(w, (v, t_v), (w, t_w)) = \sum_{d_v=1}^{H} pr(d_v | t_v) pr(d_w | t_v, d_v) \quad (3)$$

where $d = (t_w - t_v) \bmod H$, and we have marginalised out the uncertainty about $d_v$ (the uncertainty in pick-up delay).

We next address the problem of learning mobility models for individuals, which provides the probability of presence that defined the Bernoulli trial used in Equations 2 and 3.

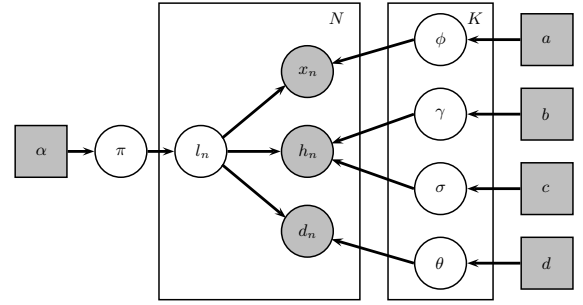

Figure 2: Graphical structure of the Dirichlet process location model, showing conditional independence between the random variables. Shaded nodes are observable and square nodes are fixed values.

## 3.2 MODEL FOR LEARNING HUMAN MOBILITY FROM CELL PHONE DATA

We now focus on the problem of getting an accurate predictive probability density of presence for any location given the participant and the time $pr(v|i, t_v)$, from which the probability of delay can be derived and used in optimisation (as described in Section 3.1). The Orange dataset consists of a set of tuples for each participant $p_i \in P$ of the form $(i, x_i, t_i)$ indicating that participant $i$ was observed near cell tower $x_i$ (discrete) at date and time $t_i$ (continuous). There are three main factors that influence the design of the model:

1. **Cell allocation noise**
   The cell tower observations provide discrete measurements on the individual's likely location. However, there may be a choice of several towers that the phone can connect to (especially in urban environments) at any single location. This allocation is decided by outside factors that we treat as noise (i.e., the network operator's optimal allocation of phones to towers). Our approach needs to isolate the human presence information in the cell tower allocation to phones and ignore other factors. This implies the need to infer the locations, each of which may be statistically associated with several cell towers.

2. **Sporadic observations**
   Since the cell tower is only recorded in this dataset when a phone call or text is made (about 7 times a day, on average) approaches that were designed to be used on continuously collected location data (e.g. eigenvectors [22, 6], variable-order Markov models [25], linear embedding [23]) are not likely to be effective (which we confirm in Section 4.2). We therefore need a method that can fill in (extrapolate from other observations) large periods of no observability.

3. **Short duration**
   The data on each individual covers a period of only 2

weeks. This, combined with the fact that each day may have only a few (or zero) observations, makes learning challenging. Overfitting is a danger when the training data (i.e., the 2 weeks of observations) contains characteristics that do not generalise to the rest of the individual's behaviour (i.e., beyond 2 weeks).

These considerations suggest the use of the Bayesian framework, which allows us to assume the existence of latent variables that abstract away from the variability of cell allocation (Factor 1), and make custom assumptions about the smoothness of location (Factor 2). Furthermore, Bayesian non-parametric methods can provide us with powerful guards against overfitting (Factor 3).

In more detail, we assume the existence of latent discrete locations, $l_n$, that are associated with each observation $(x_n, t_n)$, and correspond to places in the individual's routine life (e.g., home, work). Mixture modelling is a well established method for inferring latent discrete variables, but the standard approach requires the specification of the number of locations [1]. Therefore, we use a Dirichlet process mixture model (a non-parametric approach) that allows us to also infer the number of locations, $K$ [16]. This is important because setting $K$ too high (manually) will cause the model to overfit the data.

To address the problem of filling in large periods of missing data, we assume that behaviour is periodic, as is common in other routine mobility models [22, 23]. Specifically, we assume both weekly and daily periodicities in behaviour. In the model, we achieve this by decomposing the date/time observation $t_n$ to the discrete day of the week, $d_n$, and continuous hour of the day $h_n$. The practical implications of this choice are explored briefly in Section 4.4.

A full generative model for location observations of each individual is therefore the following:

$$\boldsymbol{\pi} \sim DP(\alpha) \quad (4)$$

for each latent location $k$ :
$$\boldsymbol{\phi_k} \sim Dir(a), \ \gamma_k \sim \mathcal{N}(b) \quad (5)$$
$$\omega_k \sim IG(c), \ \boldsymbol{\theta_k} \sim Dir(d) \quad (6)$$

for each observation $n$ :
$$l_n \sim \mathcal{M}(\boldsymbol{\pi}), \ x_n \sim \mathcal{M}(\boldsymbol{\phi_{l_n}}) \quad (7)$$
$$h_n \sim \mathcal{N}(\gamma_{l_n}, \omega_{l_n}), \ d_n \sim \mathcal{M}(\boldsymbol{\theta_{l_n}}) \quad (8)$$

where, first, distribution $\boldsymbol{\pi}$ over latent locations is drawn from a Dirichlet process (Equation 4) that defines the prior probability of each location in the dataset. Second, the four parameters to the model $\phi, \gamma, \omega, \theta$ are drawn from their prior distributions (Dirichlet, normal, inverse-gamma, and Dirichlet, respectively) in Equations 5-6 [1]. These priors were chosen for their conjugacy to the parameter distributions, making the model simpler to infer. Thirdly, for each observation, latent location $l_n$ is drawn (Equation 7), and this location defines all the observable information in the dataset ($x_n$, the cell tower, $h_n$ the continuous hour observation, and $d_n$, the day of the week). Since $x_n$ and $d_n$ are discrete observations, they can be drawn from multinomials, while $h_n$ (the continuous hour of the day) is drawn from a normal distribution with mean $\theta_{l_n}$ and variance $\omega_{l_n}$ (Equations 7-8). Defining $h_n$ in this way makes the temporal distribution smooth, allowing us to fill in periods with only a few observations. However, we sacrifice some flexibility with this assumption, i.e., it does not capture multimodalities in presence for a single location $l_n$.

The conditional independence assumptions between the random variables are visually represented in Figure 2. Direct inference of all the parameters from the data is not possible in this model, requiring us to either optimise them (i.e., variational approximation) or to perform Markov chain Monte Carlo sampling [1]. Several effective and conceptually simple Gibbs sampling schemes are available for inference with a Dirichlet process, so we used the latter approach adapted from [16]. After obtaining samples (following convergence of the Markov chain), we can find the predictive distribution for location $v$ given the entire training set $\boldsymbol{X}$ for each individual [1]:

$$pr(v|t_v, \boldsymbol{X}) = \frac{1}{R} \sum_{r=1}^{R} pr(v|t_v, \boldsymbol{M^{(r)}}) pr(\boldsymbol{M^{(r)}}|\boldsymbol{X}) \quad (9)$$

where $r$ is the index of each sample (taken after convergence), $t_v$ is the query time, $\boldsymbol{M^{(r)}}$ is the entire set of model parameters found in sample $r$, and $R$ is the total number of samples.

To test our approaches to optimisation and learning, we next apply them to the real cell tower observations.

## 4 EXPERIMENTAL RESULTS

In this section we use the real world cell tower mobility data of 50,000 people living in Ivory Coast, measured over 2 weeks, to assess the feasibility of crowdsourcing package delivery in Section 4.1. Then, using the same data, we evaluate our approach to prediction in Section 4.2, and optimisation under uncertainty in Section 4.3.

### 4.1 FEASIBILITY STUDY

To assess the feasibility of the idea of crowdsourcing package delivery, we consider three key criteria: (1) the number of participants required for acceptable geographical coverage; (2) the number of participants required in any specific delivery (since longer chains imply greater risk of loss and theft); and (3) the feasibility of delivering to rural locations, which is expected to be much harder than urban delivery. To assess these criteria, it was sufficient to consider a simpler instantiation (in this section only) of the problem we defined in Section 3.1 that takes into account the locations that each person in the participant set, $P$, visited, but does not include the temporal structure in the mobility. We consider the temporal aspect of feasibility in Section 4.3.

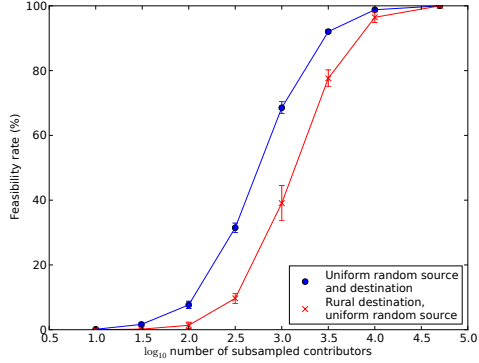

Figure 3: A plot of the percentage of randomly sampled (source,destination) delivery problems that had a solution path of any size, against the $\log_{10}$ size of the number of potential contributors.

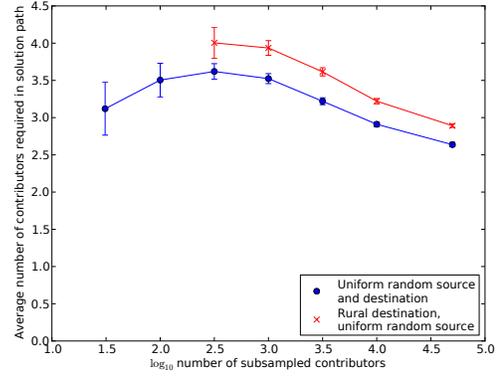

Figure 4: A plot of the average number of contributors required to each specific delivery problem (drawn from the much larger pool of potential contributors) against the $\log_{10}$ size of the potential contributors pool. N.B., a majority of rural destinations are infeasible for pool sizes of less than $10^{2.5}$, therefore we are unable to plot the line below this range.

#### 4.1.1 Criterion 1: Number of Participants Required

To assess the number of participants required for wide geographical coverage (Criterion 1), we uniformly randomly subsampled participant sets, $P'$, from the global participant set $P$ (containing 50,000 people), for a wide range of different sizes $|P'| = \{10^{0.5i} | i = 1, 2, ..., 9\}$. For each participant set, we then uniformly sampled 1,000 pairs of locations (source and destination) from $V$ representing 1,000 possible delivery problems. We consider a different (urban to rural) distribution of test locations in Section 4.1.3.

For each test location pair, we used Dijkstra's algorithm to find the shortest path (the standard algorithm can be applied to graph $\mathcal{G}$ because these is no uncertainty about the edge costs). Figure 3 shows the percentage of location pairs that were feasible (i.e., that had any path between the source and destination locations). The line with circular points shows the feasibility for uniform random source and destination locations. We see that the geographical coverage is very poor when there are fewer than $10^{2.5}$ participants. The critical range is around $10^3$, when feasibility surges with each new participant. The heavy tail in human location behaviour is one explanation for this effect, where individuals visit many locations a few times (and a few locations many times) in their daily life mobility [8]. Therefore, an acceptable geographic coverage, trading off against recruitment/administration costs, appears to be around $10^{3.5}$ participants.

#### 4.1.2 Criterion 2: Number of Participants Required for Any Given Delivery Problem

To assess the number of participants required in any given solution path (Criterion 2), we used the same subsampled participant sets as in Section 4.1.1 and plotted the length of the shortest path against the size of each subsampled participant set in Figure 4. The length of the shortest path indicates how many people are required for any specific delivery problem. The circular points are the focus for Criterion 2, where we see that the number of participants required for any solution path stays within the small range of 2 to 4. Since infeasible paths cannot be included when plotting Figure 4 (because they have unspecified numbers of contributors), the number of contributors required for specific paths initially increases with the size of the participant subset, as more paths are made feasible. However, once path feasibility (indicated in Figure 3) goes beyond 20%, the trend is as expected; having a wider pool of participants allows more efficient (i.e., shorter length) paths to be discovered. Note that, since we are not considering duration in Figure 4, the lowest cost paths in the full model may require more people. In any case, since the cost for losing the package can be fully specified by the planner, the optimal tradeoff between path length and duration can be found.

#### 4.1.3 Criterion 3: Rural Distribution

So far, we have only considered uniformly sampled source and destination test points, which favours urban locations (since there are greater numbers of cell towers in urban areas). We now consider Criterion 3 for rural feasibility, by sampling a set of delivery problems where the destinations are only rural (keeping source locations uniformly sampled, as before).

Table 1: Average $\log_e$ data likelihood (higher is better) of held out test data of 50,000 individuals. 95% confidence intervals are given.

| MODEL | LOG LIKELIHOOD |
|---|---|
| Our approach | $-5.890 \pm 0.057$ |
| First-order MM | $-6.110 \pm 0.043$ |
| VMM order 2 (Song et al. 2006) | $-6.276 \pm 0.030$ |
| VMM order 3 | $-6.347 \pm 0.033$ |
| Random | $-6.696 \pm 0.056$ |
| Finite mixture, (Cho et al. 2011) | $-9.452 \pm 0.066$ |

We ran the same analyses for Criteria 1 and 2 with rural destinations, yielding the lines with crosses in Figures 3 and 4. We conclude that restricting the destinations to be rural certainly makes the delivery problem more challenging, but it is still feasible. Now that we know that all three feasibility criteria are met, we consider the problem of learning the temporal structure in mobility to enable the minimisation of delay in delivery from source to destination nodes.

## 4.2 EVALUATION OF HUMAN MOBILITY PREDICTIONS

In this section, we evaluate our approach to predicting human mobility under considerable data sparsity, as would be typical from cell tower datasets. We split the cell tower data of 50,000 people into training and testing sets. The test set contains a single cell tower location reading from each person's data, therefore giving a test set of 50,000 data points. The rest of the data for the same individuals was used in training. To test for model quality, we looked at the logarithm of the data likelihood of each test point. We used non-informative hyperparameters $a = 1, d = 1, \alpha = 1$ for the discrete priors (see Figure 2). We used $b = (0.01, 12)$ and $c = (0.01, 3)$ for the continuous temporal priors, referring to the relative mean of precision w.r.t. the data, the mean of the prior, the degree of freedom in the precision, and the inverse mean of precision, respectively.

For comparison on the same data, we also tested two existing approaches that are considered state-of-the-art for human routine location prediction. The first is a spatio-temporal approach by Cho et al. (2011) [4], and the second method is a sequential approach by Song et al. (2006) [25] based on variable-order Markov models (VMM). In addition, we also tested a purely random model, with data likelihood $pr(x, d, h) = \frac{1}{L} \frac{1}{V} \mathcal{N}(h|\mu = 12, \sigma = 6)$, where $L$ is the total number of locations and $V = 7$ is the number of days of the week (and $(x, d, h)$ is the location, day of week, and hour of day observation as before).

The held-out data likelihood of all the approaches on the 50,000 data points can be seen in Table 1. We first note that VMM is worse than even a first-order Markov model, which is contrary to the findings of Song et al. (2006). This difference is due to the fact that the training data is very sparse, so learning higher-order dependencies causes a degradation in likelihood, even though the motivation behind fall-back (in the VMM) is to dynamically use orders appropriate to the context. Consequently, we see a further degradation as we increase the maximum order of the VMM to 3. We can also see that the model of Cho et al. (2011) performs the worst out of all the approaches. In contrast, our model outperforms all the others by at least 25% (since we are using a $\log_e$ likelihood). We believe most of this benefit comes from selecting the right number of components using the Dirichlet process (to let the data "speak for itself"). In Figure 5, we show the number of components (i.e., latent locations) found for a random subsample of 1,000 individuals, plotted against the dataset size for each individual. The number of latent locations has mean 4.1, mode 2, and standard deviation 2.4, with a heavy tail. Therefore, the bimodal assumption of Cho et al. (2011) is true for a large number of individuals in our dataset, yet, there are still many other individuals for whom their model is too complex, or not complex enough. Performance for these individuals that makes their model worse overall.

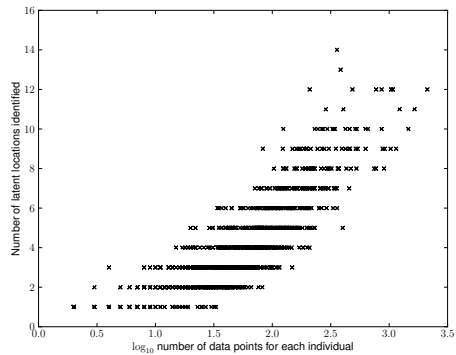

Figure 5: The number of individual latent locations identified by the Dirichlet process for 1,000 randomly selected individuals in the dataset.

## 4.3 EVALUATION OF OPTIMISATION APPROACH

We now evaluate the optimisation element of our work (i.e., which participants to ask and which intermediate locations to use). To do this, we make a few additional assumptions in light of the results we have presented so far. Firstly, since a participant pool of approximately 3,500 people is enough to get satisfactory coverage of Ivory Coast (see Section 4.1), we used participant sets of this size in our optimisation evaluation. Secondly, in order to get statistically significant results, we ran 10,000 simulations using our mobility model (given in Section 3.2) as the ground truth, since it performs best under extreme data sparsity. To evaluate the robustness of the optimisation to uncertainties in human behaviour, we considered the total delay cost of 10,000 simulations, using the approach presented in Section 3.1 (using modified policy iteration to find the optimal policy [20]). To put this result in context, we also evaluated two alternative approaches to package routing as benchmarks. The first benchmark is the naïve approach that finds the shortest path (i.e., the minimum number of contributors), but does not consider the temporal mobility habits of the participants. The second benchmark uses the findings presented in Section 3.1, but finds the path of lowest expected cost during the planning stage instead of during runtime (and is therefore a heuristic based on our MDP formulation). This results in a policy for participant selection (i.e., who to ask to deliver the package given the time slot) but a *fixed* route. We therefore expect this approach to perform worse compared to the full optimal policy, since it is not able to react to optimally to incoming delay information.

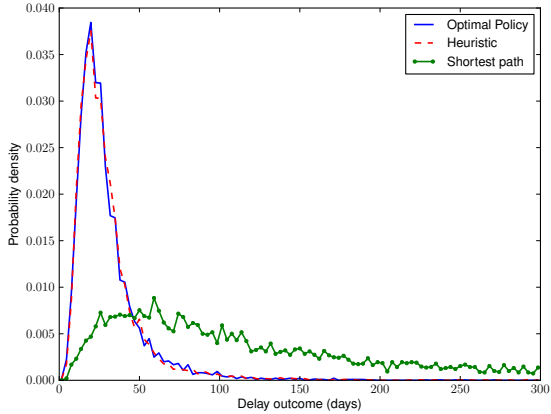

Figure 6: Probability distribution of delay cost in 10,000 simulated journeys to rural destinations using our MDP formulation (and heuristic) versus just taking the shortest path.

The results are presented in Figure 6, showing the end-to-end duration performance of our optimal policy and heuristic approaches against the shortest path benchmark. For this, we used the rural test set, defined in Section 4.1.3, with an average of 373 km between the source and destination locations. The average total duration for the optimal policy is 30.0 days, versus 161 days for the benchmark. The heuristic we based our MDP formulation on performed almost identically to the optimal policy, with an average of 30.7 days duration. Interestingly, all three distributions are heavy-tailed, which conforms to expectations from other findings about delays from human behaviour [8]. There is, therefore, an 81.3% time advantage to learning and optimising over human behaviour, and it seems that without a consideration of the mobility habits of the participants, there would be an infeasible delay. Furthermore, since our heuristic performs almost as well as the optimal policy, there appears to be little benefit to being able to dynamically (at runtime) change the next location in response to the delays observed so far.

### 4.4 DISCUSSION

To perform routing under uncertainty, we assumed that the participants would follow their normal mobility patterns when delivering packages (see Section 3.1). Clearly, additional factors could introduce further delay, including disruptions to transport and short term disruptions arising from participants' circumstances (e.g., being too busy, taking sick leave). In practical terms, most of the impact of these disruptions could be absorbed by an appropriate task assignment procedure. Specifically, after obtaining a policy from our learning and optimisation approach, the system could ask the selected participants, via automated phone text, whether they are actually willing and able to do the task. In this way, participants facing disruptions can be filtered out, limiting the introduction of unexpected delay into the route. On the other hand, some disruptions may not be known at the time of task acceptance, or some participants may simply not be honest about them. We leave this as a problem for future work (see Section 5).

Finally, in the worst case (from a routing perspective), participants may lose or steal packages. A certain amount of loss and theft is assumed even with standard delivery, and is borne as the risk of doing business, or addressed with insurance. In the crowdsourced setting, this can be taken into account by assigning a cost to each participant (either with a fixed value, or derived from a participant-specific trust evaluation framework). In whatever way the cost of trust is calculated, once obtained, it can be incorporated into the MDP as an added cost in the standard way.

### 5 CONCLUSIONS AND FUTURE WORK

In this work we studied a novel method for distribution that uses the existing mobility of local people to send packages large distances. Using data describing the real world movement patterns of 50,000 people, we addressed the technical problems associated with this method, formulating an MDP for optimisation and presenting a Bayesian non-parametric model that performs well under data sparsity. Future work could incorporate the most recent observations of participants' locations in order to respond to unexpected delays (in addition to the random variability in delay attributable to daily life mobility that we already did consider). Introducing this sequential dependence breaks the periodic feature of the predictions, making the MDP intractable again. To address this, a hybrid approach could be developed that assumes periodicity during initial planning, but which allows local refinements to the policy as up-to-the-hour information about participant mobility arrives.

# References

[1] C. M. Bishop. *Pattern recognition and machine learning*, volume 4. Springer New York, 2006.

[2] J. Chen, K. H. Low, C. K.-Y. Tan, A. Oran, P. Jaillet, J. Dolan, and G. Sukhatme. Decentralized data fusion and active sensing with mobile sensors for modeling and predicting spatiotemporal traffic phenomena. In *Proceedings of the Twenty-Eighth Conference Annual Conference on Uncertainty in Artificial Intelligence (UAI-12)*, pages 163–173, Corvallis, Oregon, 2012.

[3] M. Cherubini, M. Zhu, N. Oliver, and M. Cebrian. Exploring Social Networks as an Infrastructure for Transportation Networks. In *Presentation at the International School and Conference on Network Science (NetSci'10)*, Boston, MA, USA, 2010.

[4] E. Cho, S. A. Myers, and J. Leskovec. Friendship and mobility: user movement in location-based social networks. In *Proceedings of the 17th ACM SIGKDD international conference on Knowledge discovery and data mining*, pages 1082–1090. ACM, 2011.

[5] CIA Directorate of Intelligence. The world factbook. July 2008.

[6] N. Eagle and A. S. Pentland. Eigenbehaviors: identifying structure in routine. *Behavioral Ecology and Sociobiology*, 63(7):1057–1066, 2009.

[7] H. Gao, J. Tang, and H. Liu. Exploring social-historical ties on location-based social networks. In *6th International AAAI Conference on Weblogs and Social Media*, 2012.

[8] M. Gonzalez, C. Hidalgo, and A. Barabasi. Understanding individual human mobility patterns. *Nature*, 453(7196):779–782, June 2008.

[9] L. Hufnagel, D. Brockmann, and T. Geisel. Forecast and control of epidemics in a globalized world. *Proceedings of the National Academy of Sciences of the United States of America*, 101(42):15124–15129, 2004.

[10] B. Keller, P. von Bergen, R. Wattenhofer, and S. Welten. On the feasibility of opportunistic ad hoc music sharing. *Mobile Data Challenge by Nokia Workshop, in conjunction with International Conference on Pervasive Computing*, 2012.

[11] K. Laskey, N. Xu, and C. H. Chen. Propagation of delays in the national airspace system. In *Proceedings of the Twenty-Second Conference Annual Conference on Uncertainty in Artificial Intelligence (UAI-06)*, pages 265–272, Arlington, Virginia, 2006.

[12] J. Laurila, D. Gatica-Perez, I. Aad, J. Blom, O. Bornet, T. Do, O. Dousse, J. Eberle, and M. Miettinen. The mobile data challenge: Big data for mobile computing research. In *Mobile Data Challenge by Nokia Workshop, in conjunction with International Conference on Pervasive Computing*, Newcastle, UK, 2012.

[13] C. Liu and J. Wu. Practical routing in a cyclic mobispace. *Networking, IEEE/ACM Transactions on*, 19(2):369–382, 2011.

[14] J. McInerney, A. Rogers, and N. R. Jennings. Improving location prediction services for new users with probabilistic latent semantic analysis. *In Mobile Data Challenge by Nokia Workshop, in conjunction with International Conference on Pervasive Computing*, 2012.

[15] J. McInerney, J. Zheng, A. Rogers, and N. R. Jennings. Modelling heterogeneous location habits in human populations for location prediction under data sparsity. In *International Joint Conference on Pervasive and Ubiquitous Computing (UbiComp 2013)*, in press.

[16] R. M. Neal. Markov chain sampling methods for dirichlet process mixture models. *Journal of computational and graphical statistics*, 9(2):249–265, 2000.

[17] E. Nikolova, M. Brand, and D. R. Karger. Optimal route planning under uncertainty. In *Proceedings of International Conference on Automated Planning and Scheduling*, 2006.

[18] E. Nikolova and D. R. Karger. Route planning under uncertainty: The canadian traveller problem. In *Proc. AAAI*, pages 969–974, 2008.

[19] G. Pickard, I. Rahwan, W. Pan, M. Cebrian, R. Crane, A. Madan, and A. Pentland. Time critical social mobilization: The darpa network challenge winning strategy. 2010.

[20] M. L. Puterman. *Markov decision processes: Discrete stochastic dynamic programming*. John Wiley & Sons, Inc., 1994.

[21] I. Rahwan, S. Dsouza, A. Rutherford, V. Naroditskiy, J. McInerney, M. Venanzi, N. Jennings, and M. Cebrian. Global manhunt pushes the limits of social mobilization. *IEEE Computer*, 46(4):68–75, 2013.

[22] A. Sadilek and J. Krumm. Far out: Predicting long-term human mobility. In *Twenty-Sixth AAAI Conference on Artificial Intelligence*, 2012.

[23] S. Scellato, M. Musolesi, C. Mascolo, V. Latora, and A. Campbell. Nextplace: a spatio-temporal prediction framework for pervasive systems. In *Pervasive Computing*, pages 152–169, San Francisco, CA, USA, 2011. Springer.

[24] J. Scott, A. J. Brush, J. Krumm, B. Meyers, M. Hazas, S. Hodges, and N. Villar. PreHeat: controlling home heating using occupancy prediction. In *Proceedings of the 13th international conference on Ubiquitous computing (UbiComp 2011)*, pages 281–290, Beijing, China, 2011.

[25] L. Song, D. Kotz, R. Jain, and X. He. Evaluating next-cell predictors with extensive wi-fi mobility data. *IEEE Transactions on Mobile Computing*, 5(12):1633–1649, 2006.

[26] E. Stevens-Navarro, Y. Lin, and V. W. Wong. An mdp-based vertical handoff decision algorithm for heterogeneous wireless networks. *Vehicular Technology, IEEE Transactions on*, 57(2):1243–1254, 2008.

[27] V. Vukadinović, Ó. R. Helgason, and G. Karlsson. A mobility model for pedestrian content distribution. In *Proceedings of the 2nd International Conference on Simulation Tools and Techniques*, page 93. ICST (Institute for Computer Sciences, Social-Informatics and Telecommunications Engineering), 2009.

[28] A. Wesolowski, N. Eagle, A. J. Tatem, D. L. Smith, A. M. Noor, R. W. Snow, and C. O. Buckee. Quantifying the impact of human mobility on malaria. *Science*, 338(6104):267–270, 2012.

[29] J. H. Wu and R. Givan. Feature-discovering approximate value iteration methods. *Abstraction, Reformulation and Approximation*, pages 901–901, 2005.